\documentclass[dvips,11pt,twoside]{article}
\usepackage{natbib}
\usepackage{url}
\usepackage[dvips]{graphicx}

\renewcommand{\figurename}{{\small Fig.}}
\renewcommand{\tablename}{{\small Table}}

\makeatletter
\long\def\@makefigurecaption#1#2{\footnotesize%
 \vskip\abovecaptionskip
\setbox\@tempboxa\hbox{\textbf{#1}#2}%
  \ifdim \wd\@tempboxa >\hsize
    \unhbox\@tempboxa\par
  \else
    \hbox to\hsize{\hfil\box\@tempboxa\hfil}%
  \fi}
\long\def\@makecaption#1#2{%
  \vskip\abovecaptionskip
  \sbox\@tempboxa{#1. #2}%
  \ifdim \wd\@tempboxa >\hsize
    {\small #1. #2}\par 
  \else
    \global \@minipagefalse  
    \hb@xt@\hsize{\hfil\box\@tempboxa\hfil}%
  \fi
  \vskip\belowcaptionskip}
\def\fnum@figure{\figurename\thefigure}
\def\fnum@table{\tablename\nobreakspace\thetable}
\renewenvironment{table}
               {\let\@makecaption\@maketablecaption%
               \let\@floatboxreset\@tableboxreset\@float{table}}
               {\end@float}
\renewenvironment{table*}
               {\let\@makecaption\@maketablecaption%
               \let\@floatboxreset\@tableboxreset\@dblfloat{table}}
               {\end@dblfloat}
\def\@tableboxreset{%
        \reset@font%
        \centering\footnotesize%
        \def\arraystretch{1.0}
        \@setnobreak%
        \@setminipage%
}
\newdimen\tablewidth \tablewidth\textwidth
\newdimen\saved@tablewidth \saved@tablewidth\textwidth
\long\def\@maketablecaption#1#2{%
 \begingroup%
    \footnotesize%
    \global\setbox\@tempboxa\hbox{#2}%
 \endgroup%
 \centering%
 \ifdim \wd\@tempboxa>\tablewidth %
    \parbox[t]{\tablewidth}{\footnotesize \hfill#1\hfill\vphantom{Ay}\par #2\vphantom{Ay}\par}%
 \else
    {\small #1} \par \hbox 
    to\hsize{\hfill\box\@tempboxa\vphantom{Ay}\hfill}
 \fi%
 \global\saved@tablewidth\tablewidth%
 \global\tablewidth\hsize\vskip\belowcaptionskip}

\newdimen\mpindentsz\mpindentsz=\the\parindent 
\def\mpindent{\leavevmode\hbox to \mpindentsz{\phantom{a}\hss}}
\usepackage{times}
\normalfont
\usepackage[T1]{fontenc}

\makeatletter
\def\ps@myownheadings{%
   \def\@oddfoot{\small 1560-4292/08/\$17.00 \copyright \ 2016 -- Springer 
   Press and the authors. All rights reserved\hfil\mbox{}}
    \def\@evenfoot{\small 1560-4292/08/\$17.00 \copyright \ 2016 -- 
    Springer and the authors. All rights reserved\hfil\mbox{}} 
    \def\@evenhead{\parbox{12cm}{\small International Journal of 
    Artificial Intelligence in Education 18 (2016) xxx 
    -- xxx \hfil\thepage
    \protect\\ Springer}\hfil\mbox{}}%
    \def\@oddhead{\parbox{\textwidth}{\small International Journal of 
    Artificial Intelligence in Education 18 (2016) xxx 
    -- xxx  \hfill\small\thepage
    \protect\\ Springer}\hfil\mbox{}}%
    }
   \def\ps@myheadings{%
   \let\@oddfoot\@empty\let\@evenfoot\@empty
   \def\@evenhead{\small\thepage\hfil\slshape\leftmark\hfil\mbox{}}%
   \def\@oddhead{\small\mbox{}\hfil{\slshape\rightmark}\hfil\thepage}%
    \let\@mkboth\@gobbletwo
    \let\sectionmark\@gobble
    \let\subsectionmark\@gobble
    }
\makeatother

\begin{document}\thispagestyle{myownheadings}

\mbox{}\\
\begin{flushleft}
    \noindent{\LARGE\bf
Automated assessment of non-native learner essays: Investigating the role of linguistic features}
\end{flushleft}

\vskip 1cm \noindent\large{\bf
Sowmya Vajjala,}
{\it 
Iowa State University, USA\\
sowmya@iastate.edu\\
}

\vskip 22pt 
\noindent\normalsize
\begin{small} 
{\bf Abstract.} Automatic essay scoring (AES) refers to the process of scoring free text responses to given prompts, considering human grader scores as the gold standard. Writing such essays is an essential component of many language and aptitude exams. Hence, AES became an active and established area of research, and there are many proprietary systems used in real life applications today. However, not much is known about which specific linguistic features are useful for prediction and how much of this is consistent across datasets. This article addresses that by exploring the role of various linguistic features in automatic essay scoring using two publicly available datasets of non-native English essays written in test taking scenarios. The linguistic properties are modeled by encoding lexical, syntactic, discourse and error types of learner language in the feature set. Predictive models are then developed using these features on both datasets and the most predictive features are compared. While the results show that the feature set used results in good predictive models with both datasets, the question "what are the most predictive features?" has a different answer for each dataset. 
\end{small}
\pagestyle{myheadings}

\vskip 22pt 
\noindent\normalsize
\begin{small}
\end{small}

\newpage
\vskip 22pt
\section{\large\bf Introduction}                                              
People learn a foreign language for several reasons such as living in a new country or studying in a foreign language. In many cases, they also take exams that certify their language proficiency based on some standardized scale. Automated Essay Scoring (AES) refers to the process of automatically predicting the grade for a free form essay written by a learner in response to some prompt. This is commonly viewed as one way to assess the writing proficiency of learners, typically non-native speakers of a language. Producing such an essay is also one of the components of many high stakes exams like GRE$^{\tiny \textregistered}$, TOEFL$^{\tiny \textregistered}$ and GMAT$^{\tiny \textregistered}$. Several AES systems are already being used in real world applications along with human graders. Along with such high stakes testing scenarios, AES could also be useful in placement testing at language teaching institutes, to suggest the appropriate level language class to a learner. Owing to this relevance to different language assessment scenarios,  AES has been widely studied by educational technology researchers. While most of the published research on AES has been on proprietary systems, recent availability of publicly accessible learner corpora facilitated comparable and replicable research on second language (L2) proficiency assessment \citep{Yannakoudakis.Briscoe.ea-11,Nedungadi.Raj-14}. 

One non-test taking application of automatic analysis of writing is in providing real-time feedback to the writers on their language use, highlighting their mistakes and suggesting ways for them to improve their writing. This has long been used as a part of word processing software (e.g., Microsoft Word), and more recently, tools such as grammarly.com, turnitin.com, e-rater$^{\tiny \textregistered}$ and WriteToLearn$^{\tiny \texttrademark}$are being used in such scenarios for analysis of student writing. A related area of research, but not specific to language use, is automatic content assessment, typically for scoring short answers to questions \citep{Burrows.Gurevych.ea-15}. Considering that many people enrolling in Massive Open Online Courses (MOOCs) are non-native English speakers, such automated methods could also be useful in providing feedback for them in scenarios where they have to write long essays as a part of evaluation for some of the courses, especially in arts and humanities. Thus, AES can also be useful in both these scenarios to assess the language used by the learners. 

Despite this long history of research and real-life applications, very little is known about what linguistic features are good predictors of writing proficiency. One reason could be that most of the published research on the topic used only one data source to develop and evaluate their models. The generalizability of the linguistic features described in one approach to another dataset is also not explored much in existing research. In this background, this article explores the following research questions:
\begin{itemize}
\item Can we build good predictive models for automatically evaluating learner essays, based on  a broad range of linguistic features?
\item What features contribute the most to the accuracy of automatic essay scoring systems?
\item Are the features generalizable to other datasets that have a different grading scheme and different target learner groups? That is, are the most predictive features the same across datasets, or is the predictive power of a feature specific to a corpus?
\item What role does the native language of the writer play in their second language writing proficiency prediction?
\end{itemize}

These questions are addressed using two publicly available second language writing corpora containing text responses written in response to some prompt. Several linguistic features that encode word, POS, syntax, discourse level information, and error features were developed using free, open-source NLP software, to build predictive models of L2 writing. While some of these features were used earlier in AES studies, some are newly introduced in this paper. The role of native language (L1) in L2 essay scoring has been documented in past research on AES \citep[e.g.,][]{Chodorow.Burstein-04} but has not been explored in great detail. This paper also takes a step in this direction and studies the role of L1 in predicting L2 proficiency. Thus, this research has a dual aim of understanding the linguistic properties of L2 writing proficiency (according to human graders) as well as developing predictive models for writing assessment. To summarize, the primary contributions of this paper are as follows: 
\begin{itemize}
\item Predictive models of L2 writing for AES are constructed using a wide range of linguistic features, several of which have not been used in this context before.
\item The question of the generalizability of linguistic features used for AES, which was not studied in detail earlier, is given importance, by exploring two datasets for this task. Existing research generally reported results only with a single dataset.
\item The role of a learner's native language as a feature in predictive modeling of L2 writing proficiency was not explored much in past research. This paper takes a first step in this direction.
\end{itemize} 

The rest of the paper is organized as follows: First, a collection of related work on this topic is discussed putting the current research in context. The two sections that follow describe the methodological framework in terms of the corpora used and the features studied. The next section describes the experimental setup, evaluation methods and results of the experiments. The paper concludes with a summary of the results, a discussion on the implications of the results and pointers to future work.

\section{\large\bf Related Work}         
\label{sec:related}
Automated Essay Scoring (AES) has been an active area of research for about four decades now and several assessment systems such as Intelligent Essay Grader \citep{Landauer.Lahm.ea-03}, Project Essay Grade \citep{Page-03}, E-Rater \citep{Burstein-03,Attali.Burstein-06} and IntelliMetric \citep{Elliot-03} are being used in real-world applications as a companion to human graders.  AES systems are typically developed using human scorer judgments as the gold standard training data to build automatic prediction models using textual features. Hence, the purpose of existing AES systems is to emulate a human grader. Contemporary AES systems use a wide range of features to build their predictive models, ranging from superficial measures like word length and sentence length to sophisticated natural language processing based approaches. The features employed to measure written language cover aspects such as: grammatical and orthographic correctness, language quality, lexical diversity and fluency \citep{Dikli-06}. The definitions of individual aspects are typically based on human expert judgments. Some of the commonly used features in AES systems are: essay length, word length, counts of various parts of speech, syntactic structure, discourse structure and errors. Distributional semantics based models such as Latent Semantic Analysis were also used in AES systems in the past \citep[][]{Landauer.Lahm.ea-03}. 

Despite a wide spread of AES systems, there is not a lot of published work on what are the different linguistic features that contribute to AES accuracy. One reason could be that the AES systems are proprietary software. Recent years saw the release of a few public datasets that can be used for developing AES models \citep[e.g.,][]{Randall.Groom-09,Yannakoudakis.Briscoe.ea-11,Kaggle-12,Blanchard.Tetreault.ea-13}. \cite{Yannakoudakis.Briscoe.ea-11} released a corpus of learner essays written for Cambridge First Certificate in English (FCE) exam, and conducted several experiments to do predictive modeling of human scores on these essays. This was extended in \cite{Yannakoudakis.Briscoe-12} who focused exclusively on how discourse coherence can be modeled for learner essays. More recently, \cite{Crossley.Kyle.ea-14} studied the relationship between linguistic properties and TOEFL scores using several linguistic indices based on Coh-Metrix \citep{Graesser.McNamara.ea-12} and Writing Assessment Tool \citep{Crossley.Roscoe.ea-13}. They used a corpus of 480 essays scored on a scale of 1--5 for this study. In another study, \cite{Crossley.Kyle.ea-15} studied the role of various linguistic features in AES using a corpus of 997 persuasive essays written by students at four grade levels. In \cite{Crossley.Kyle.ea-16}, they explored the relationship between cohesion features and writing quality using a corpus of 171 English learner texts. Apart from predictive modeling based approaches, there are also studies that focused on identifying distinctive linguistic features between proficiency levels \citep*[e.g.,][]{Tono-00,Lu-10,Lu-12,Vyatkina-12} and L1 backgrounds \citep{Lu.Ai-15}.

Most of the research in AES has been related to English writing owing to its widespread use and the availability of more learner corpora and language processing software for the language. However, the past half-decade saw the emergence of AES research in non-English (primarily European) languages. \cite{Ostling.Smolentzov.ea-13} developed an AES approach for detecting Swedish language proficiency using a corpus of high-school level exams conducted nationwide in Sweden. \cite{Hancke.Meurers-13} described a proficiency classification approach for a publicly accessible dataset of German learner essays, based on the CEFR \citep{CouncilofEurope-01} scale used in Europe. \cite{Vajjala.Loo-13,Vajjala.Loo-14} developed an approach for automatically predicting Estonian learner proficiency on the CEFR scale, also based on a public dataset. In developing the features, all the above-mentioned approaches relied on the specific properties of the language (e.g., morphology) along with features generally used in English. However, to our knowledge, AES systems developed for non-English languages have not been put to widespread use in any real life application the way English AES systems are being used (yet). 

Despite this background of  a wide body of work, there is not much of published research on what specific linguistic properties of learner writing contribute to receiving a better proficiency score. Having such knowledge is not only useful for the improvement of AES systems, but also for systems that provide feedback to learners, such as e-rater$^{\tiny \textregistered}$ and WriteToLearn$^{\tiny \texttrademark}$. Exploring the relation between such linguistic features and proficiency scores is one of the issues addressed in this paper. While some of the features employed in this paper overlap with the features described in previous studies and all these papers essentially talk about the same underlying research issues, the current paper differs in both the methodologies employed, and the overall features used. The second issue addressed in this paper relates to the generalizability of the linguistic features, which is addressed by multi-corpus evaluation using two large publicly available corpora.

There has been some recent research modeling task independence in AES \citep{Zesch.Wojatzki.ea-15,Phandi.Chai.ea-15}, and they used another publicly available dataset of scored essays \citep{Kaggle-12}. The research described in this paper differs from this strand of research primarily in two aspects: Firstly, they focus on modeling the differences between tasks, and adaptation of a model trained on one task data to another, and not on the specific features that contribute to the AES as such. Secondly, their experiments make use of only one corpus. Finally, one of the specific research questions in our case is to investigate native language influence on L2 proficiency prediction. This was not also modeled in these two articles as the native language background of learners in Kaggle dataset is not known.

Generalizability of a machine learning approach to essay scoring can be studied in two ways: multi-corpus study (training multiple models, each with a different training data) and a cross-corpus study  (training with one corpus and testing it using another). We study feature generalizability using a multi-corpus study setup in this paper. Doing a multi-corpus study using two corpora that come from very different sources, and graded by human graders using different norms makes it possible for us to compare what features work for which corpus specifically. Since we do not know the exact guidelines for grading for individual corpora, it will not be appropriate to do a cross-corpus study in this scenario. Further, a cross-corpus study will only tell us about the generalizability of a model built using one dataset (with specific feature weights) on another. But a multi-corpus study will help us address our specific research question about feature generalizability i.e., what features are useful to develop predictive models across multiple data sources. 

In terms of research on publicly available corpora, the current work can compare closely to \cite{Yannakoudakis.Briscoe.ea-11} and \cite{Yannakoudakis.Briscoe-12}, who worked on the First Certificate of English corpus, which is one of the corpora used in this paper. In contrast with the pairwise-ranking approach used in their work, our model uses regression. While we model similar aspects of text as both these papers in our approach, the feature set described in this paper contains fewer, but denser features. It will be shown that our models achieve a comparable performance with the reported results on this dataset. Thus, compared to existing work on AES, this paper reports experiments with a new corpus, uses some new features that were not used in this context before, and compares the feature performance with more than one corpus. To our knowledge, this is the first multi-corpus study of automatic essay scoring task.              

\section{\Large\bf Methods: Corpora}
\label{sec:corpora}
The experiments were conducted using two publicly accessible corpora that have human scorer annotations of language proficiency for essays written by non-native English speakers. They are described below:

\subsection{\bf TOEFL11 Corpus}
Our first corpus for the experiments reported in this paper is the TOEFL11 corpus of non-native English \citep{Blanchard.Tetreault.ea-13}. This is a collection of essays written by TOEFL iBT\textsuperscript{\textregistered} test takers in 2006-2007 in response to the \textit{independent writing} task in the test. The learners responded to one of the 8 prompts, each of which differed in the topic of the question and not in the task itself. The entire corpus consists of 12100 essays written by learners with 11 L1 backgrounds and belonging to three proficiencies (low, medium, high). These proficiency categories were a result of collapsed TOEFL scores that were originally on a scale of 1--6. The essays were written in response to eight prompts. The first version of this corpus that was released during the First Native Language Identification Shared Task \citep{Blanchard.Tetreault.ea-13} is used in this paper for the experiments. This version had essays sorted by the native language of the learners and had 900 texts per L1. However, the proficiency distribution was not even (5366 text samples for medium, 3464 for high, 1069 for low). 

In the initial classification experiments, it was observed that this created a strong bias for the medium proficiency, which resulted in poor precision and recall for the other two proficiency categories. Hence, a sample of 1069 texts per category was selected, using the SpreadSubSample method implemented in WEKA toolkit \citep{Hall.Frank.ea-09} to train balanced classifiers described in later sections of this paper.\footnote{The file ids of the text files used in this experiment can be shared for replication studies.} We will refer to this corpus as \textsc{TOEFL11Subset} for the rest of this paper. Table~\ref{tab:corpusdesc} shows average number of tokens and sentences per text for the three proficiency categories in our corpus. 

\begin{table}[htb!]
\caption{\textsc{TOEFL11Subset} (1069 texts per category)}
\begin{tabular}{lll}
\hline \textbf{Proficiency}&\textbf{avg. tokens}&\textbf{avg. sentences}\\
 low&297.2&11.7\\
medium&378.3&16.1\\
high&421.9&17.8\\
\hline
\end{tabular}
\label{tab:corpusdesc}
\end{table}

This corpus has been primarily used for the native language identification shared task and its derivatives, and its usefulness for automated scoring of essays has not been explored much in research. The only available research that used this corpus for such a purpose is \cite{Horbach.Poitz.ea-15}, and they rely exclusively on trigram frequency features, compared to the rich feature set described in this paper.  

\subsection{\bf FCE Corpus}
Our second corpus is the First Certificate of English (FCE) corpus that was publicly released by \cite{Yannakoudakis.Briscoe.ea-11}. They released a corpus of First Certificate of English exam transcripts, which is a subset of the larger Cambridge Learner Corpus (CLC). The corpus contains texts produced by takers of English as Second or Other Language (ESOL) exams, and the associated scores on a scale of 1--40. These texts are typically letters and personal essays written in response to given prompts.

Since this corpus has a numeric scale, and with a much broader scale range than \textsc{TOEFL11} corpus which had only three categories, we did not look into balancing the corpus for all the scores, as that would have resulted in a very small corpus rendering it useless for machine learning purposes. However, we used the same train-test setup as described in previous research that used this corpus (1141 texts from year 2000 for training, 97 texts from year 2001 for testing) for comparability. This will enable us to do a direct comparison with other reported research on this corpus. We will refer to these as \textsc{FCE-Train} and \textsc{FCE-Test} respectively for the rest of this paper. 

It has to be noted that compared to \textsc{TOEFL11} corpus where all prompts elicited same form of response (writing an analysis of agreement or disagreement with a given statement), the prompts in \textsc{FCE} corpus asked for different forms of response such as a letter, a report, a short story and so on. 

As mentioned earlier, since the native language of the writer is one of the features considered in the models described below, Kaggle corpus \citep{Kaggle-12} was not considered in this study, as it does not provide this information about the writer's background. Additionally, Kaggle corpus also involved aggressive preprocessing for anonymization, with operations such as removing all named entities \citep{Blanchard.Tetreault.ea-13}. This would also have affected some of our features, especially the discourse features. 

\section{\Large\bf Methods: Features}
\label{sec:features}
A broad range of features covering different aspects of linguistic modeling appropriate for learner texts were developed in this paper. There is no publicly available documentation on how exactly are the human scorers assigning scores to individual essays, apart from the general guidelines. Typically, AES research described features that can be automatically extracted, while encoding different linguistic aspects related to written language. The feature choices described in this article also follow a similar direction, while also considering what we know from Second Language Acquisition (SLA) about learner writing. L2 writing proficiency in SLA research has been discussed in the context of the notions of Complexity, Accuracy and Fluency (CAF) \citep{Housen.Kuiken-09}. This paper describes automatically extractable features that can encode these three aspects at different levels of language processing. For complexity, features that study the lexical richness and syntactic complexity in SLA, and features from readability assessment research are used. For accuracy, features modeling errors made by learners are extracted. For fluency, features typically used to model discourse coherence of texts in computational linguistics were used. 

In terms of the linguistic aspects encoded, the features are classified into 6 groups: word level, parts of speech, syntactic characteristics, discourse properties, errors, and others. All the underlying linguistic representations (tags, parses, coreference information etc.,) for the feature calculations described below are obtained from the Stanford Parser \citep{Socher.Bauer.ea-13} and the Stanford CoreNLP package \citep{Manning.Surdeanu.ea-14}.\footnote{The code for feature extraction is hosted at: \url{https://bitbucket.org/nishkalavallabhi/ijaiedpapercode/}}

\subsection{\bf Word Level Features}
This group of five features consists of the measures of lexical diversity typically used in first and second language acquisition studies as a measure of the diversity in a learner's vocabulary. While there are many measures of lexical diversity, we grouped some of the measures that do not rely on any linguistic representation other than words alone into this feature group. This group consists of four variations of type-token ratio (TTR) with the formulae described in \cite{Lu-12}, which are TTR ($num. types/num. tokens$), Corrected TTR ($num. types/\sqrt{(2*num. tokens)}$), Root TTR ($num. types/\sqrt{(num. tokens)}$), and Bilog TTR ($Log$ $num.types/Log$ $num.tokens$) and a fifth measure called Measure of Textual Lexical Diversity (MTLD) as described in \cite{McCarthy.Jarvis-10}, which measures the average length of continuous text sequence that maintains the TTR above a threshold of 0.72. Since these features capture the diversity of language use, we can hypothesize that they will be useful for L2 proficiency prediction. While these features were used to do a corpus based analysis of their correlation with L2 proficiency \citep{Lu-12}, they were not used in the context of predicting modeling of writing proficiency before, to our knowledge. This feature group will be referred to as \textsc{WordFeatures} in the rest of the paper. 

Since all the five formulae primarily depend on the two variables -total number of tokens and total number of unique tokens in a text, there is a high degree of correlation between most of these \textbf{five} features. However, since our purpose is to develop predictive models and predictions need not necessarily be affected by such multicollinearity, all the features, including the correlated ones, are kept in the machine learning models. Feature selection will be performed in the later part of this paper, to select the most predictive features for models. It will also be shown later how a seemingly small change in the operationalization of the feature can change the direction of relationship.  

\subsection{\bf POS Tag Density}
This group consists of \textbf{27} features that are calculated based on the percentage of various POS tags in the text and some of the measures of lexical variation used in SLA research to assess learner texts \citet{Lu-12}. We took a subset of features described in this paper that rely on ratios between POS tags (Noun, verb, adjective, adverb and modifier variation formulas). All these POS tag based features will be referred to as \textsc{POSFeatures} in the rest of the paper. They are listed in Table~\ref{tab:posfeatureslist}, grouped according to the motivations for these features. All features in this group, except the word variation features are calculated with the number of words in the text as the denominator. The variation features follow the formulae described in \citet{Lu-12}. While the lexical variation features have been used to study their correlations with L2 writing proficiency, they were not used in the context of predictive modeling before, to our knowledge. Some version of POS tag based features are used in almost all reported AES research, although the exact operationalizations may differ between this article and other research. 

\begin{table}[thb!]
    \caption{{\small Features based on POS Tags}}
    \label{tab:posfeatureslist}
\begin{center}
    \begin{tabular}{lll}
    \textbf{Lex. Variation features from \cite{Lu-12}} & \textbf{General POS tags} & \textbf{Verb tags}\\ \hline
    POS\_adjectiveVariation & POS\_numNouns&POS\_numVerbsVBD\\
POS\_adverbVariation & POS\_numProperNouns&POS\_numVerbsVBG\\
POS\_correctedVerbVariation1&POS\_numPronouns&POS\_numVerbsVBN\\
POS\_modifierVariation&POS\_numPerPronouns&POS\_numVerbsVBP\\
POS\_nounVar&POS\_numAdjectives&POS\_numVerbsVBZ\\
POS\_squaredVerbVar1&POS\_numAdverbs&POS\_numModalVerbs\\
POS\_verbVar1&POS\_numConjunctions&\\
POS\_verbVar2&POS\_numInterjections&\\
POS\_numLexicalWords&POS\_numDeterminers\\
&POS\_numPrepositions&\\
	&POS\_numVerbs&\\
	&POS\_numWhPronouns&\\ \hline
	\end{tabular}
\end{center}
\end{table}

\subsection{\bf Syntactic Complexity}
This group consists of 28 features extracted from the syntactic parse trees of sentences in learner essays. 14 of the features were used in the past to measure syntactic complexity in second language writing and its relation to writing proficiency \citep{Lu-10,Lu-11}. These features are implemented based on the descriptions in \cite{Lu-10} and using Tregex tree pattern matching tool \citep{Levy.Andrew-06} with syntactic parse trees, for extracting specific patterns. Although these 14 features were used to do a statistical analysis of second language writing in the past, they were not used in training any automated writing assessment models. The remaining 14 features estimate the average number and size of various phrasal groups (NP, VP, PP, WH-phrases) and measures of parse tree height (average height, number of subtrees, constituents, SBARs etc.) per sentence. This group will be referred to as \textsc{SynFeatures} in the rest of the paper. They are listed in Table~\ref{tab:synfeatureslist}. To our knowledge, these specific operationalizations of features (e.g., PPs per sentence, RRCs per sentence) have not been reported in AES research before.

\begin{table}[thb!]
    \caption{{\small Features based on syntactic parses}}
    \label{tab:synfeatureslist}
\begin{center}
    \begin{tabular}{ll}
     \textbf{SLA features from \cite{Lu-10}} & \textbf{Other Syntactic Features} \\ \hline
SYN\_avgSentenceLength&SYN\_avgParseTreeHeightPerSen\\
SYN\_MeanLengthofClauses&SYN\_numSentences\\
SYN\_MeanLengthofTunits&SYN\_numConstitutentsPerSen\\
SYN\_ComplexNominalsPerClause&SYN\_numConjPPerSen\\
SYN\_CNPerTunit&SYN\_avgNPSize\\
SYN\_ComplexTunitRatio&SYN\_numNPsPerSen\\
SYN\_CoordinatePhrasesPerClause&SYN\_numPPSize\\
SYN\_CoordPerTunit&SYN\_numPPsPerSen\\
SYN\_DependentClauseRatio&SYN\_numRRCsPerSen\\
SYN\_DependentClausesPerTunit&SYN\_numSBARsPerSen\\
SYN\_TunitComplexityRatio&SYN\_numSubtreesPerSen\\
SYN\_VPPerTunit&SYN\_numVPSize\\
SYN\_numTunitsPerSen&SYN\_numVPsPerSen\\
SYN\_numClausesPerSen&SYN\_WhPhrasesPerSen\\ \hline
    	\end{tabular}
\end{center}
\end{table}
    
\subsection{Discourse Properties}
Text coherence is one of the scoring criteria used in essay scoring \citep{Burstein.Tetreault.ea-10}. Coherence also relates to the \textit{fluency} aspect of the CAF framework in SLA research, which makes it a relevant aspect to consider in L2 writing proficiency prediction models. However, there is no single method to model coherence and different approaches have been proposed in computational linguistics research so far. Hence, several discourse features are considered in this paper, that encode coherence with different levels of linguistic analysis, based on existing research in natural language processing. 

At a very shallow level, there are a total of eight word overlap features - content word, noun, stem, and argument overlap at  local (between adjacent sentences) and global (between any two sentences in a text) levels. The implementation of these overlap features is based on the descriptions in the Coh-Metrix tool \citep{Graesser.McNamara.ea-12} documentation. This group of \textbf{eight} features will be referred to as \textsc{Disc-Overlap} in the rest of the paper. These features have been used in some of the recent writing assessment studies that used Coh-Metrix tool \citep{Crossley.Kyle.ea-16}. 

At the level of part of speech, referential expressions like pronouns and articles are known to be some of the indicators of text cohesion. Hence, referential expression features based on the descriptions of \cite{Todirascu.Francois.ea-13} (who used them to measure coherence and cohesion for readability assessment of texts) were implemented. This group consists of \textbf{10} features that model the use of definite articles and pronouns (all, personal, possessive) per word and per sentence, and pronouns and proper nouns per noun. These will be referred to as \textsc{Disc-RefEx} in the rest of the paper. To our knowledge, referential expression features were not explicitly studied in the context of automatic proficiency assessment before. 

Discourse connectives (words such as: \textit{and, although, but, however}, etc.,) that connect parts of a text are frequently used as a measure of text coherence in literature. Previous corpus research also emphasized the importance of studying connector usage in learner texts \citep{Granger.Tyson-96}. Some of the existing tools like Coh-Metrix\footnote{\url{http://cohmetrix.com/}} provide features that calculate the occurrence of various types of connectives in text based on word lists (e.g., causal, temporal, contrastive, etc.,). However, not all connective words are used always as discourse connectives in a text. \cite{Pitler.Nenkova-09} described an approach to disambiguate the discourse versus non-discourse usage of connectives based on the syntactic parse trees. Along with this, they also provide a sense classification for the discourse usage of connectives (four senses: Expansion, Contingency, Comparison, Temporal). This method is available as a downloadable discourse connectives tagger,\footnote{\url{http://www.cis.upenn.edu/~nlp/software/discourse.html}} which takes parse trees of sentences in a text as input, and tags it with discourse annotations. This tool was used to calculate \textbf{seven} connectives based features: the number of discourse and non-discourse connectives, all connectives, and the number of occurrences of each of the four senses, per sentence. This group will be referred to as \textsc{Disc-Conn} in this paper. To our knowledge, these features too have not been used for automatic proficiency assessment before. 

\cite{Barzilay.Lapata-08} introduced a model of text coherence based on the concept of an entity grid, where the various mentions of an entity in a text are labeled with their syntactic roles (subject, object, neither, and absent). These roles are then used to build a grid of entity transitions which capture how an entity changes across different sentences in a text. \cite{Burstein.Tetreault.ea-10,Yannakoudakis.Briscoe-12} used entity grid based features for automated writing assessment before and they were among the more predictive feature groups in \cite{Yannakoudakis.Briscoe-12}. The current paper uses \textbf{16} features based on the transitions between the four syntactic roles for the entities (subject to subject, subject to object, subject to absent, subject to neither, and similar features for object, other, and neither respectively) and 4 additional features that calculate the average number of entities per sentence and per text, number of unique entities per text and average number of words per entity as additional features, as used in some of the earlier research in analyzing text coherence for readability assessment \citep{Feng.Jansche.ea-10}. This group of features is referred to as \textsc{Disc-Entities} in this paper. 

Co-reference chains refer to the chains of references to the same entity in a piece of text, as we progress from one sentence to another. They are a useful way to infer coherence in texts and were used in readability assessment research \citep{Schejter-15} as a measure of text coherence. In this paper, \textbf{8} co-reference chain features based on noun phrases, nouns and pronouns and determiner usage in the essay were used. These features will be referred to as \textsc{Disc-Chains} in this paper. To our knowledge, these features have not been used for AES earlier and this is the first article to report their utility for this task. Table~\ref{tab:refchains} lists the coreference chain features.

\begin{table}[htb!]
\caption{Reference Chain Features}
\begin{tabular}{ll}
\hline
average length of a reference chain\\
proportion of personal pronouns in a reference chain\\
proportion of demonstrative pronouns in a reference chain\\
proportion of reflexive pronouns in a reference chain\\
proportion of proper nouns in a reference chain\\
proportion of possessive determiners in a reference chain\\
proportion of demonstrative determiners in a reference chain\\
proportion of indefinite Noun Phrases (NP) in a reference chain\\
proportion of definite NPs in a reference chain\\ \hline
\end{tabular}
\label{tab:refchains}
\end{table}

\subsection{\bf Errors}
Errors are one of the most intuitive features to use in assessing the proficiency of a writer. Some of the existing research on this topic used large language models with word and POS unigrams to model learner errors \citep[e.g.,][]{Yannakoudakis.Briscoe.ea-11}). In this paper, an open source rule based spelling and grammar check tool called LanguageTool\footnote{\url{https://languagetool.org/}} was used to calculate spelling and grammar features. The following \textbf{four} error features were extracted for each text: average number of spelling errors, non-spelling errors and all errors per sentence, and the percentage of spelling errors for all errors in a text. This group of features will be referred to as \textsc{ErrorFeatures} in this paper.\footnote{The primary reason for  choosing tool is the fact that it is under active development, and provides both spelling and grammar check in one place, eliminating the need to develop those modules for this study. We are not aware of any other such off-the-shelf library for spelling and grammar check.}

\subsection{\bf Document Length} 
Document length is one of the features used in all AES systems and is known to be a strong predictor of proficiency. Hence we included it as a feature, encoding it as the number of tokens in the document. This will be referred to as \textsc{docLen} for the rest of this paper. It can be argued that document length may not be a useful feature to have in language exams, as they usually have a word limit. However, it has been used in all research and production AES systems, and is known to correlate strongly with proficiency and hence has been used in this paper as one of the features.

\subsection{\bf Others: Prompt and L1}
Apart from all the above-mentioned features, prompt (the question for which the learners responded with answers) and the native language of the learner (L1) are considered as additional (categorical) features for both datasets. 

Linguistic properties of writing prompts were also shown to influence the student writing \citep{Crossley.Varner.ea-13}. Since there is prompt information for both datasets, this was included in the feature set. However, it has to be noted that prompt was used only as a categorical feature without including any features that account for prompt specific vocabulary in the learner essays.

While L1 was not directly considered as a feature in the L2 essay scoring approaches before, the possible influence of L1 on L2 proficiency prediction was discussed in previous research \citep{Chodorow.Burstein-04}. Recently, \cite{Lu.Ai-15} showed that there are significant differences in the syntactic complexity of L2 English writing produced by college level students with different L1 backgrounds. L1 specific differences in terms of linguistic complexity were also seen in the feature selection experiments with \textsc{TOEFL11 dataset} in the recent past using readability assessment features \cite[Chapter~7]{Vajjala-15}. On a related note, two recent studies \citep{Tetreault.Blanchard.ea-12,Kyle.Crossley.ea-15} showed that L2 proficiency influenced the prediction accuracy of native language identification. In this background, we can hypothesize that L1 can be useful as a predictive feature for proficiency classification. 

Both \textsc{TOEFL11Subset} and FCE have the L1 information for texts. So, this was used as a feature for training models with both datasets. Though all the L1s in \textsc{TOEFL11Subset} corpus had sufficient representation across all proficiencies, \textsc{FCE} corpus had a large imbalance in L1 representation across the score scale, as it had learners with more diverse L1 backgrounds compared to \textsc{TOEFLSubset}. Nevertheless, it was used as one of the features in the prediction models. However, we can hypothesize that this may not be as useful with \textsc{FCE} dataset. 

Word and POS ngrams were typically used in other AES systems in the past to model learner language and compare it with "native" English. They were also used as a means to measure error rates in learner language \citep{Yannakoudakis.Briscoe.ea-11}. However, they were not used in this paper for the following reasons:
\begin{itemize}
\item The features described above were all chosen to model clear linguistic properties, and are hence dense in nature. Ngram features are relatively sparse features, which can become difficult to interpret, compared to the dense features, in terms of what they are encoding linguistically. 
\item Ngram features can also be topic  and vocabulary specific, and may not be very informative if we model across prompts, which differ in lexical choices.
\item One major reason for using ngram features in previous research was to model errors in learner language compared to native English writing. However, in this paper, error features are designed based on LanguageTool, which internally models errors based on linguistic rules and ngrams.
\end{itemize}

\section{\Large\bf Experiments and Results}
\label{sec:results}
The experiments reported in this paper primarily belong to two categories: development of predictive models, and a study of the most useful features for predictive models. The overall approach for conducting the experiments and interpreting the results is described below: 
\subsection{\bf Approach}
\label{subsec:approach}
The scores in the two datasets used in this paper are designed differently. While the \textsc{TOEFLSubset} has three categories, the \textsc{FCE} dataset has a numeric score ranging from 1--40. Hence, the following two approaches were adapted to enable a comparison between them:
\begin{enumerate}
\item Treat \textsc{TOEFLSubset} prediction as classification and \textsc{FCE} dataset as regression according to how the datasets are designed, and then compare what features are more predictive for both datasets, using a common feature selection approach which will work for both classification and regression methods.
\item Since both datasets are ordinal in nature, we can convert \textsc{TOEFLSubset} too into a numeric scale (low = 1, medium = 2, high = 3) and compare both datasets using a common prediction algorithm. While it is  possible to create a discretized version of \textsc{FCE} too, that will not consider the fact that both datasets are ordinal in nature. Hence, that conversion is not performed in this paper. 
\end{enumerate}

WEKA toolkit \citep{Hall.Frank.ea-09} was used to train and test our prediction models. For \textsc{TOEFLSubset}, the classification models were evaluated in terms of classification accuracy in a 10 fold cross validation setting. For \textsc{FCE}, the performance of the models was assessed in terms of Pearson correlation and Mean Absolute Error (MAE) performance on \textsc{FCE-Test} set. While 10 fold CV may give us a better estimate about the stability of the prediction model, there is some published research on \textsc{FCE} dataset with the exact train-test split as the one used in this paper. Thus, it is possible to do a direct comparison of the results on this dataset with existing research. So we focus on the results on test set for this corpus, and use 10 fold CV to compare between \textsc{TOEFLSubset} and \textsc{FCE} datasets. For training the models, multiple classification and regression algorithms were explored. Sequential Minimal Optimization (SMO), which is a support vector machine variant, worked well among these algorithms. Further, WEKA has both a classification and regression variant for this algorithm. So, SMO was used to train all the models described below, with normalized feature vectors and linear kernel. 

\subsection{\bf Classification with \textsc{TOEFL11Subset}}
As mentioned earlier, both classification and regression experiments were performed on this dataset. The first classification model with this corpus consisted of all the features described in the Features section. This model achieved a classification accuracy of 73.2\%. Table~\ref{tab:confusion} shows the confusion matrix for the classifications. It can be seen that the largest confusion exists between low--medium and medium--high and hardly any overlap exists between low and high proficiencies. Consequently, medium proficiency is the most difficult to predict. This is not surprising, and is to be expected, if we remember that language proficiency is a continuum. Further, the confusion matrix works as an evidence that the classification model is actually learning the continuum of relationship between low/medium/high proficiencies.

\begin{table}[htb!]
\caption{\textsc{TOEFL11Subset} Classification Summary}
\label{tab:confusion}
\begin{tabular}{l|lll}
\hline \textbf{Classified as -->}&\textbf{low}&\textbf{medium} & \textbf{high}\\ 
\hline low&\textbf{881}&180&8\\
medium&157&\textbf{666}&246\\
high&4&264&\textbf{801}\\
\hline
\end{tabular}
\end{table}

Removing prompt and L1 as features resulted in less than 0.5\% decrease in classification accuracy. So, it looked as if there is no effect of these two features on the corpus at least when we use the full feature set. With this number as our comparison point for further experiments, we studied the impact of individual feature groups for this task. For each feature group, a small ablation test is reported as well, removing prompt and L1 as features. Table~\ref{tab:toeflsummary} shows the results for the feature groups. Random baseline refers to the accuracy in the case where the classifier always picks only one class irrespective of the input. In our case, since all the three classes are represented equally, this becomes 33\%. Document length alone as a feature resulted in a 64.3\% classification accuracy, but created a skew towards low and high proficiencies, thereby affecting the precision and recall for medium proficiency. 

\begin{table}[htb!]
\caption{\textsc{TOEFL11Subset} Feature Group Performance}
\begin{tabular}{llccc}
 \hline \textbf{Feature Group} & \textbf{Num. Features} & \textbf{Accuracy}  & \textbf{Accuracy without prompt}  & \textbf{Accuracy without prompt and L1} \\
 \hline Random Baseline & --  &33.0\% & 33.0\% & 33.0\% \\
 \hline \textsc{docLen} & 1  &66.3\% & 66.3\% & 64.3\% \\
\hline \textsc{Word}&  5 &67.4\%&68.0\% & 66.9\%\\ 
 \textsc{POS} & 27 & 68.2\% &67.8\% &66.4\% \\
 \textsc{Syn} & 28  &63.6\% & 63.3\% & 61.1\% \\
 \textsc{Disc-All} & 49  &61.4\% & 61.8\% & 59.2\%\\
\hline \textsc{ Disc-Overlap} &  8 &56.8\%&56.8\% &\textbf{52.4\%} \\
 \textsc{  Disc-RefEx} & 10  &48.8\%& 48.8\%& \textbf{42.0\%}\\
 \textsc{  Disc-Conn} & 7 &48.7\% & 48.7\%& \textbf{36.0\%}\\
 \textsc{  Disc-Entities} & 16 &49.0\% &49.0\% &\textbf{40.5\%} \\
 \textsc{  Disc-Chains} &  8 &48.7\%&48.7\% & \textbf{39.4\%}\\
\hline \textsc{Error} & 4 &51.0\% &51.3\% &48.2\%  \\ \hline
 All Features & 114 +2 (prompt, L1)  &73.2\% & 73.1\%& 73.0\%\\ \hline 
 \end{tabular}
\label{tab:toeflsummary}
\end{table}

Looking at the performance of feature groups, it can be noticed that the features that do not need much linguistic analysis (docLen, POS, Word) performed better than more linguistically demanding features. Despite this, the model with all features outperformed the next best feature group (POS) by 5\%. Error features performed the poorest as a stand alone group, achieving only 51\% accuracy. However, it should be remembered that the current approach only considers 4 features based on errors, and only considers two broad classes of errors (spelling and non-spelling). The classification accuracies were also in general lower for the discourse feature groups compared to other groups. While prompt seemed to have very little influence over classification accuracies, dropping L1 resulted in a huge drop in performance for all the discourse feature groups (marked in bold in Table~\ref{tab:toeflsummary}). This drop is large as 12\% for \textsc{Disc-Conn} features and around 5-10\% for other discourse feature groups. However, when all the discourse features are combined, removing L1 feature has resulted in only a 2\% reduction in accuracy. Though we do not have any hypothesis about the reasons for this L1 influence on specific discourse features, this seems to indicate that there are differences in the usage of discourse markers by people with different L1 backgrounds, across the proficiency levels. This could be an interesting direction to explore in future.  

\paragraph{Feature Selection:} To understand how many features we really need to reach the performance of all features put together, a range of feature selection methods provided in WEKA were explored - which used information gain, gain ratio, one R classifier, correlation and distance from other instances as the characteristics for feature selection. Twenty to thirty features were sufficient to reach the classification accuracy of 72-73\% with all the feature selection methods explored, which was the accuracy with all features put together. In all cases, this smaller subset included features from all the five major feature groups (word, pos, syntax, discourse, errors). Since a detailed comparison of top features from individual feature selection methods is beyond the scope of this article, we will focus on top-10 features using one method, ReliefF \citep{Robnik-Sikonja.Kononenko-97}, which can be applied for both categorical and numeric class values.

\subsection{\bf Regression with \textsc{TOEFL11Subset}}
\textsc{TOEFL11Subset} corpus is released as a categorical corpus. However, proficiency ratings are ordinal in nature and are not categories independent from each other. Further, to enable a comparison with FCE which has a numeric scale, \textsc{TOEFLSubset} was modeled as regression, using SMOReg algorithm. The model achieved a Pearson correlation of 0.8 and a Mean Absolute Error (MAE) of 0.4 with actual proficiency values. The Pearson correlation of different feature groups with proficiency is summarized in Table~\ref{tab:toeflregsummary}, which follows the same pattern as the classification model in terms of the feature group performance and the influence of L1 on discourse features (marked in bold in Table~\ref{tab:toeflregsummary}).

\begin{table}[htb!]
\caption{\textsc{TOEFL11Subset} Feature Group Performance}
\begin{tabular}{llcc}
 \hline \textbf{Feature Group} & \textbf{Pearson Corr.}  & \textbf{Corr. without prompt}  & \textbf{Corr. prompt and L1} \\
 \hline \textsc{docLen} &  0.67 &  0.67 &  0.65 \\
\hline \textsc{Word} & 0.70 & 0.69 & 0.67 \\ 
 \textsc{POS} & 0.74 & 0.74 & 0.72 \\
 \textsc{Syn} &  0.62 & 0.62  & 0.56  \\
 \textsc{Disc-All} & 0.66  & 0.65  &  0.61\\
 \textsc{Disc-Overlap} & 0.59 & 0.59 & 0.49  \\
 \textsc{Disc-RefEx} & 0.40 & 0.40 & \textbf{0.24} \\
 \textsc{Disc-Conn} &  0.40 & 0.40 & \textbf{0.16} \\
 \textsc{Disc-Entities} & 0.40 & 0.40 & \textbf{0.13}\\
 \textsc{Disc-Chains} & 0.40& 0.40 &\textbf{0.15}\\
 \textsc{Error} &  0.55 &  0.55& 0.40 \\ \hline
 All Features & 0.80 & 0.79 & 0.80 \\ \hline 
 \end{tabular}
\label{tab:toeflregsummary} 
\end{table}

The ten most predictive features according to ReliefF feature selection method for this dataset are shown in Table~\ref{tab:toefltop10}, excluding prompt and native language. The algorithm selects attributes by repeatedly sampling instances in a dataset and comparing the value of the selected attribute for the nearest instances to the current instance.  
\nocite{Kononenko-94}

\begin{table}[htb!]
\caption{\textsc{TOEFL11Subset} Most Predictive Features}
\begin{tabular}{ll}
\hline \textbf{Feature} & \textbf{Group}  \\
 Document length & -- \\
 Corrected Type Token Ratio& \textsc{Word} \\
 Root Type Token Ratio& \textsc{Word} \\
  percentage of spelling errors& \textsc{Errors} \\
 num. sentences per text& \textsc{Syn} \\
 squared verb variation & \textsc{POS} \\
 corrected verb variation& \textsc{POS} \\
 global stem overlap& \textsc{Disc} \\
 global argument overlap& \textsc{Disc} \\
  global noun overlap& \textsc{Disc} \\
\hline \end{tabular}
\label{tab:toefltop10}
\end{table}

Since ReliefF selects features independently, it does not capture the fact that some of the features can be correlated, as in the case of type token ratio variants and verb variation variants in Table~\ref{tab:toefltop10}. Despite that shortcoming, it can be observed from the table that there are features belonging to all the five major categories of features represented in this paper. Additionally, an important observation is: despite the fact that the top 10 features have features from all groups, they all rely on relatively shallow representations of language. The discourse features in this group are all word overlap features, which do not require a deep discourse analysis of the text. These 10 features together achieved a correlation of 0.76 in a 10 fold cross validation experiment, which is very close to what was achieved with the whole feature set. However, this model still had features related to different aspects of language, though that need not necessarily mean a deeper modeling of linguistic structures. 

While feature selection methods that choose individual best features and rank them is a good approach to follow, it has to be noted that the ReliefF selection method used above is independent of the learning algorithm, and is more of a dataset characteristic. What will give a better understanding of how the features work together in a model is a comparison of the features with highest positive and negative weights in the regression model. This can also facilitate a comparison with another model's feature weights. Table ~\ref{tab:toeflweights} below lists the five most discriminative features with positive weights and five with negative weights in the TOEFL11 regression model. It has to be noted that the weights refer to the feature weights in the context of the overall model, and not their individual weights independent of other features.

\begin{table}[htb!]
\caption{\textsc{TOEFL11Subset} Feature Weights}
\begin{tabular}{lll|lll}
\hline \multicolumn{3} {l} \textbf{high positive weight} & \multicolumn{3} {|l} \textbf{high negative weight} \\
\hline weight & feature & feature group & weight & feature & feature group \\
\hline +1.31 & document length & -- & -1.51 & num. non-spelling errors & \textsc{Errors}\\
\hline +0.78 & num. temporal connectives & \textsc{Disc}  & -0.85 & num. proper nouns & \textsc{POS}\\
\hline +0.57 & num. interjections & \textsc{POS} & -0.72  &num. spelling errors  & \textsc{Errors}\\
\hline +0.49 & pronouns to nouns ratio & \textsc{Disc} & -0.72 & num. personal pronouns & \textsc{Disc} \\
\hline +0.48 & num. T-units per sentence & \textsc{Syn} & -0.60 & num. non-discourse connectives & \textsc{Disc}\\
\hline
 \end{tabular}
\label{tab:toeflweights} 
\end{table}

It is not surprising to see that two error features are among the features with high negative weights, which implies that a higher number of errors results in a low proficiency prediction. Longer texts result in a higher proficiency score, as is evidenced by document length feature having a high positive weight. POS and Discourse features are seen in both positively and negatively discriminating features, and there is a syntactic feature from SLA (num. T-units) among the most predictive positive weight features, which is consistent with the results from analysis of L2 writing in SLA research \cite{Lu-10}. Number of non-discourse connectives got a negative weight for this model, whereas number of temporal connectives got a positive weight. This implies that the sense of usage of connective words, and whether they are used as connectives at all or as normal words in the text plays a role in L2 writing proficiency assessment.

\subsection{\bf With FCE corpus}
Now, we turn to the question of how much these observations hold when we use a different corpus. Such an experiment will enable us to compare between the two datasets and observe what features work on both datasets and what work with only one dataset. To explore this direction, a regression model was trained with all the features using \textsc{FCE-Train} as the training set in a 10-fold CV setting. The model achieved a Pearson correlation of 0.63 and a Mean Absolute Error (MAE) of 3.4 with the actual scores. On testing with \textsc{FCE-Test} (containing 97 texts), the model had a correlation of 0.64 and a MAE of 3.6. 

The same train-test setup was used in \cite{Yannakoudakis.Briscoe.ea-11} and \cite{Yannakoudakis.Briscoe-12}. In the first study, they reported a highest correlation of 0.74 between the predicted and actual scores. In a second study with coherence features, they reported a highest correlation of 0.75 between actual and predicted scores. As we are testing on the same group of texts as previous research, it is possible to compare the correlations in terms of statistical significance. The comparison was performed using \textit{cocor} correlation comparison library in R \citep{Diedenhofen.Musch-15} which compares two correlations using a variety of statistical tests and there was no significant difference between the correlations reported in \cite{Yannakoudakis.Briscoe.ea-11} and what was reported in this paper (0.64) with any of the tests. However, the result from the second study \citep{Yannakoudakis.Briscoe-12} is significantly better (p $<$ 0.05) than the current model, which can perhaps be attributed to the presence of additional coherence features in their model (e.g., those based on incremental semantic analysis). Since these studies did not report a measure for error margins, we cannot compare the MAE values we got with existing work. Nevertheless, these results show that the feature set shows comparable performance with other reported results on this dataset. 

Like with the experiments using \textsc{TOEFL11Subset}, the next step with this corpus too is to study which features contribute to a good prediction. Table~\ref{tab:fceSummary} shows a summary of model performance with feature groups on the \textsc{FCE-Test} data. For each feature group, we built three models - one with all features from the group, one excluding prompt and one excluding native language. However, in all the cases, removing prompt and native language did not have a significant effect on the results. So, only the results with all features (= features of the group+prompt+nl) are reported in the table.

\begin{table}[htb!]
\caption{Feature Group Performance on \textsc{FCE-Test}}
\begin{tabular}{lcc}
\hline \textbf{Feature Group} & \textbf{Pearson Correlation}  & \textbf{MAE} \\ \hline
\textsc{docLen}&0.31&4.6\\
 \textsc{Word}&0.29&4.8\\ 
 \textsc{POS} &0.49 & 4.4\\
 \textsc{Syn} &0.44 &4.4\\
 \textsc{Disc-All} & 0.55&4.0 \\
 \textsc{Disc-Overlap} &0.19 &4.7\\
 \textsc{Disc-RefEx} &0.34 & 4.8\\
 \textsc{Disc-Conn} & 0.04& 5.0\\
 \textsc{Disc-Entities} & 0.09 &5.1\\
 \textsc{Disc-Chains} & 0.43&4.3 \\
 \textsc{Error} & 0.46&4.0 \\
 \textsc{All Features} &0.64&3.6 \\
\hline \end{tabular}
\label{tab:fceSummary}
\end{table}

Discourse features as a group seem to be the single best performing feature group for this data set, followed by \textsc{POS} and \textsc{Error} features. Within the discourse features, lexical chain features that require coreference resolution are the best performing features, while all other discourse features except referential expressions performed poorly on this corpus. In general, compared to the \textsc{TOEFL11Subset}, stand alone feature groups perform poorly compared to the entire feature set put together. Document length, which was very predictive of the performance for \textsc{TOEFL11} dataset did not do particularly better in comparison with other features for FCE dataset. 

To understand which individual features had the most predictive power on this dataset, we again used the ReliefF algorithm. Table~\ref{tab:fcetop10} shows the ranked list of 10 best features and their categories.
 \begin{table}[htb!]
 \caption{Best Features on \textsc{FCE-Train} dataset}
\begin{tabular}{ll}
\hline \textbf{Feature} & \textbf{Group}  \\
\hline Average Length of a reference chain & \textsc{Disc}\\
 All Errors (spelling and non-spelling) & \textsc{Error}\\
 proportion of possessive determiners in a ref. chain& \textsc{Disc}\\
 Non-spelling errors & \textsc{Error}\\
 Corrected Type Token Ratio& Word \\
 Root Type Token Ratio& Word \\
 document length \\
 proportion of demonstrative pronouns in ref. chain & \textsc{Disc}\\
 num. conjunctions/num. words & \textsc{POS}\\
 Co-ordinate phrases per T-unit &  \textsc{Syn}\\
 \end{tabular}
\label{tab:fcetop10}
\end{table}
\\ Discourse features are the most represented group in the top 10 features, followed by word and error features. It is interesting to note that compared to the best features on \textsc{TOEFL11Subset}, there is an overlap only with respect to the lexical diversity features (TTR and CTTR). With respect to the error features, while percentage of spelling errors was the most predictive feature or \textsc{TOEFL11Subset}, non-spelling errors and all errors were the most predictive for \textsc{FCE}. Additionally, co-reference chain features, which did not figure in the most predictive features for \textsc{TOEFL11Subset} were among the best features for this corpus. 

As with \textsc{TOEFL} model, Table~\ref{tab:fceweights} below shows the individual features that had the most positive and negative weights for this model.
 
 \begin{table}[htb!]
\caption{\textsc{FCE} Feature Weights}
\begin{tabular}{lll|lll}
\hline \multicolumn{3} {l} \textbf{high positive weight} & \multicolumn{3} {|l} \textbf{high negative weight} \\
\hline weight & feature & feature group & weight & feature & feature group \\
\hline +0.31 & complex nominals per T-unit & \textsc{Syn} & -0.35 & TTR & \textsc{Word}\\
\hline +0.27 & num. pronouns per noun & \textsc{Disc}  & -0.35 & document length & --\\
\hline +0.25 & proper nouns to nouns ratio & \textsc{Disc} & -0.34  & num spelling and non-spelling errors & \textsc{Errors}\\
\hline +0.24 & average sentence length & \textsc{Syn} & -0.32 & num. proper nouns& \textsc{POS} \\
\hline +0.22 & corrected TTR & \textsc{Word} & -0.28 & T-unit complexity ratio & \textsc{Syn}\\
\hline
 \end{tabular}
\label{tab:fceweights} 
\end{table}

An interesting aspect of this list of features is that, document length is a negative valued predictor, while sentence length is a positive predictor for FCE texts. Compared to \textsc{TOEFL} table (Table~\ref{tab:toeflweights}), there are a few differences regarding the importance of features in the model. There is no overlap among the positively weighted features of the two models, except for pronouns to nouns ratio, which is a discourse feature modeling referential expressions. Among the negatively weighted features, while spelling and non-spelling errors as separate features carried more weight in the TOEFL model, the error feature that combines both of them had more weight for the FCE model. Number of proper nouns is the feature that occurs commonly between TOEFL and FCE datasets with a negative weight. It is interesting to see the two variations of type-token ratio - TTR (types/tokens) and corrected TTR (types/$\sqrt {2*tokens}$) appear on either side of the weight table for FCE dataset. Further, while proper nouns per sentence, POS feature, had a negative weight, proper nouns to nouns ratio, a referential discourse feature, had a positive weight in the model. Thus, the exact operationalization of the features can change the direction of the relationship with the score variable. This shows the usefulness of considering closely related features in the prediction model. It would not have been possible to understand these relations otherwise. 

A common criticism of using a wide range of features in essay scoring has been that several features are merely proxies of document length. While this may be true to some extent, the results so far stress the importance of considering diverse features for constructing the predictive models. Further, the dependency on document length can also be an artifact of the data. One way to verify if that is the case is to look at the partial correlations of the features controlling for document length. This was done using \textit{ppcor} R package \citep{Kim-15}. For example, consider the feature corrected TTR. In the TOEFL corpus, the feature had a correlation of 0.6 with proficiency, but that dropped to a partial correlation of 0.3 when controlled for document length. However, in FCE corpus, the feature had a correlation of 0.36, and the partial correlation when controlling for document length was 0.25, which is not a  drastic drop compared to TOEFL corpus. Similarly, consider another feature, \textit{average length of the reference chain}, which is among the more predictive features of FCE corpus. One would expect that this feature, by definition, would highly vary with document length. That is, short documents cannot have very long reference chains owing to their length. So, a text with a long reference chain can be expected to have larger number of words. Thus, any correlation of this feature with proficiency can possibly be due to the document length. In TOEFL corpus, this feature had a very low correlation of -0.02 with proficiency, and the partial correlation controlling for document length is -0.13. However, with FCE corpus, this feature had a correlation of -0.43 with proficiency, but the partial correlation after controlling for document length was still -0.42. Thus, the dependence on document length of a certain feature could be an artifact of the corpus too, in some cases.  

So far, from these experiments, the differences between the two datasets can be summarized as follows:
\begin{enumerate}
\item The two datasets differ from each other not only in terms of features that individually have a better predictive power with the essay score, but also in terms of features assigned high or low weights in a prediction model. 
\item Features requiring deeper linguistic analysis (reference chain and syntactic complexity features) are more predictive in \textsc{FCE} dataset whereas the most predictive features of \textsc{TOEFL11Subset} were dominated by features requiring relatively shallow linguistic analysis, even when covering discourse aspects (e.g., word overlap).
\item While document length was the most predictive feature in TOEFL dataset, it had a relatively low correlation with essay score in FCE dataset.
\item Removing native language as a feature drastically reduced the classification accuracy for discourse feature groups in \textsc{TOEFL11Subset}. However this feature did not have any influence in the \textsc{FCE} dataset. 
\end{enumerate}

One reason for native language not having any effect in \textsc{FCE} could be due to the size of the dataset, in which there are learners with too many native language backgrounds, and each language does not have enough representation for a machine learning model to learn. Further, some languages from the training set are not seen in the test set at all. So, perhaps this dataset is not necessarily suitable to analyze L1 influence on L2. Other reason for having different features among the top predictors for both datasets may lie in the fact that they are texts of a different genre of writing. \textsc{FCE} texts contained more of narrative/conversational writings compared to TOEFL essays where the prompts always asked the writers to take a stance on some topic. The fact that \textsc{FCE} had a wider scale compared to \textsc{TOEFL11} could be another factor. Further, since the grading rubrics used by both exams, and the process of grading are different, we can also infer that language proficiency means different things to different grading schemes. So, it would not be a wise idea to use a model trained on one dataset with another, in such a case. However, a comparison with another corpus will still be useful in providing such insights about the nature of language proficiency, while also providing us an idea of how to interpret the output prediction of a given AES system.  

Thus, the primary conclusions from these experiments so far are:
\begin{itemize}
\item There may not be single best group of features that work for all datasets. One dataset may require features that demand deeper linguistic processing, and another dataset may not.
\item Considering a larger group of features covering various linguistic aspects and levels of processing (instead of having a large group of features covering one aspect) however gave us better predictive models in both cases. 
\item The relationship of native language to L2 proficiency, especially with respect to the discourse features, needs to be explored further in future work.
\end{itemize}

\section{\Large\bf Conclusions}  
This paper dealt with the issue of what linguistic features are more predictive for automatic scoring of English essays. To answer this question, several automatically extracted linguistic features that encode different aspects of language were explored, using free and open language processing software and resources. While some of these features were used for this task before, several of them were newly used in this paper. The usefulness of these features was studied by building predictive models using two public datasets - \textsc{TOEFL11Subset} and \textsc{FCE}. This makes the paper the first multi-corpus study for this task, and based on non-proprietary datasets. The paper's conclusions in terms of the original research questions from the start of the paper are as follows:
\begin{itemize}
\item \textit{Can we build good predictive models for automatically evaluating learner essays, based on  a broad range of linguistic features?}
  \\ \textsc{TOEFL11Subset}, the best model achieved a prediction accuracy of 73\% for classifying between three proficiencies (low, medium and high), using all the features. Almost all of the classification errors fell into differentiating between low--medium and medium--high with less than 1\% of the classification errors happening between low--high. When modeled as regression, the best correlation with proficiency of 0.8 was achieved with a model using all features. With \textsc{FCE}, the best model achieved a correlation of 0.64 and a Mean Absolute Error of 3.6, on the test data. This performance is comparable to other results reported on this corpus. In this backdrop, we can conclude that the feature set can result in good models for scoring the English essays of non-native speakers, with the currently available NLP approaches for pre-processing and feature extraction. It has to be noted, however, that these conclusions are restricted to what is possible with currently available language processing methods, and the predictive power of the features is also dependent on the accuracy of the underlying tools. 

\item \textit{What features contribute the most to the accuracy of automatic essay scoring systems?}
\\ In \textsc{TOEFL11Subset}, document length was the best single predictor, and word and POS level features turned out to be the best feature groups. Looking at the features individually, there is at least one feature for each of our five major feature groups among the 10 most predictive features, and the features primarily relied on shallow linguistic representations of the language aspect. For example, shallow discourse features such as word overlap were more predictive for this dataset compared to deeper ones like reference chains. Error features were among those that got high negative weight in this model, and number of T-units per sentence, which is known in SLA research to correlate with language proficiency, was among the features with high positive weight. Removing native language as a feature resulted in a drop in classification accuracy for models using discourse features. So, native language of the author seems to be a useful feature to consider while modeling the discourse of L2 writing.

In the \textsc{FCE} dataset, discourse features were the best performing group followed by POS and Error features. Native language and prompt did not have any effects on this dataset, and non-spelling errors were better predictors than spelling errors. While document length was among the best predictors, it was not as influential as in \textsc{TOEFLSubset}. In terms of the individual features, this dataset also had features from all five groups among the best features. However, features that relied on deeper linguistic modeling (such as reference chains) had more weight compared to other features. Error features and SLA based syntactic features were among the features with high positive or negative weight in this dataset too. 

One important aspect to remember is that the description of feature contributions here is limited to specifically features that are useful in the development of predictive models for scoring essays in terms of the language proficiency, and does not necessarily mean features that generally capture the broader construct of language proficiency, which involves aspects beyond essay writing. 

\item \textit{Are the features generalizable to other datasets, used with a different grading scheme and different learner groups? Are the most predictive features same across the datasets, or is the predictive power of a feature specific to a corpus?}
\\ From the current experiments, the features do not seem to be completely generalizable across datasets. There is some overlap between most predictive features of both dataset, but not to the extent that we can claim generalizability. These results lead us to conclude that there may not be a universal feature set that predicts proficiency across all L2 writing datasets. But, the best models are always achieved by modeling multiple language dimensions in the features, and by considering deeper processing of texts. While we require features that encode multiple aspects of language from word level to discourse level, what features are more important depends on the nature of the dataset used. This could also directly relate to the fact that the two datasets originate from different exams, with probably different guidelines for grading and evaluation.

\item \textit{What role does the native language of the writer play in their second language writing?}
\\ Native language of the author as a feature was an important predictor for some feature groups in one dataset, but the effect was not seen in the other dataset. In the \textsc{TOEFL11Subset} experiment, there was a clear drop in performance of discourse feature based models when native language was removed from the feature set. While the effect was not seen in FCE dataset, it could also be because of insufficient representation of all native languages in the data. Nevertheless, the results on TOEFL dataset lead us to a conclusion that native language of the author can be a useful predictor for second language scoring models. Further analysis is needed in this direction to explore the influence of native language on the discourse properties of learner essays. 
\end{itemize}

\subsection{\bf Future Work} 

One missing aspect in the models described here is that the actual word and phrase usage in the corpus has not been modeled at all in the features. While it is possible that the features currently considered indirectly capture such information, using features based on topical word/n-gram frequency measures, modeling prompt specificity explicitly should be pursued in future work. Another useful direction to take for handling this aspect will be to consider task independent features \citep[e.g.,][]{Zesch.Wojatzki.ea-15} for modeling learner language. 

Prompt has only been considered as a categorical feature in this paper, whereas it clearly needs to be better modeled in the context of recent research that showed how topic affects L2 writing quality \citep{Yang.Lu.ea-15}. Modeling if the essay actually answers the question asked in the prompt is also an important aspect which needs to be addressed in future, which will also make AES a useful tool in scoring responses to assess the comprehension of learners. Adding the question relevance in these models will make it useful for large scale content assessment, in addition to assessing language form. The influence of native language on discourse features performance in the \textsc{TOEFL11Subset} is an interesting observation and needs to be investigated in better detail to understand the role of native language in learners' written discourse. Looking at particular error types instead of two broad categories (spelling, non-spelling) as done in the current article may result in a better modeling of errors in L2 writing. Further, a qualitative analysis of the learner essays in terms of the most discriminative features and why they differ in terms of positive and negative weights with different datasets can give us better insights into the development of automated writing feedback systems. Finally, having a better understanding of the exact grading criteria used by human graders can result in a more theoretically grounded feature design in future. 

\section{\large\bf Acknowledgements}
I would like to thank all the anonymous reviewers and the editors of the issue for their useful comments, which greatly helped improve this paper from its first version. I also thank Eyal Schejter for sharing his code to extract the coreference chain and entity density features. 

\bibliography{naaclhlt2016}
\bibliographystyle{natbib}
\end{document}